\documentclass[manuscript]{acmart}

\author{Nurit Cohen-Inger}
\orcid{0009-0009-3227-3090}
\email{cohening@post.bgu.ac.il}
\author{Lior Rokach}
\orcid{0000-0002-6956-3341}
\email{liorrk@post.bgu.ac.il}
\author{Bracha Shapira}
\orcid{0000-0003-4943-9324}
\email{bshapira@bgu.ac.il}
\author{Seffi Cohen}
\orcid{0000-0002-1135-0079}
\email{seffi@post.bgu.ac.il}

\affiliation{
  \institution{Ben Gurion University}
  \city{Beer Sheva}
  \country{Israel}
  \postcode{8410501}
}

\AtBeginDocument{%
  }

\setcopyright{acmlicensed}
\copyrightyear{2025}
\acmYear{2025}
\acmDOI{XXXXXXX.XXXXXXX}

\acmConference[FAccT '25]{Proceedings of the ACM Conference}{June 23--26,
  2025}{Athens, Greece}
\acmISBN{978-1-4503-XXXX-X/24/06}

\usepackage{amsmath,amsfonts}
\usepackage{algorithm}
\usepackage{algpseudocode}
\usepackage{multirow}
\usepackage{graphicx}
\usepackage{hyperref}
\usepackage{booktabs}
\usepackage{newfloat}
\usepackage{listings}
\usepackage{natbib}
\usepackage{url}

\begin{document}

\title{FairTTTS: A Tree Test Time Simulation Method for Fairness-Aware Classification}

\begin{abstract}
Algorithmic decision-making has become deeply ingrained in many domains, yet biases in machine learning models can still produce discriminatory outcomes, often harming unprivileged groups. Achieving fair classification is inherently challenging, requiring a careful balance between predictive performance and ethical considerations. We present FairTTTS, a novel post-processing bias mitigation method inspired by the Tree Test Time Simulation (TTTS) method. Originally developed to enhance accuracy and robustness against adversarial inputs through probabilistic decision-path adjustments, TTTS serves as the foundation for FairTTTS. By building on this accuracy-enhancing technique, FairTTTS mitigates bias and improves predictive performance. FairTTTS uses a distance-based heuristic to adjust decisions at protected attribute nodes, ensuring fairness for unprivileged samples. This fairness-oriented adjustment occurs as a post-processing step, allowing FairTTTS to be applied to pre-trained models, diverse datasets, and various fairness metrics without retraining. Extensive evaluation on seven benchmark datasets shows that FairTTTS outperforms traditional methods in fairness improvement, achieving a 20.96\% average increase over the baseline compared to 18.78\% for related work, and further enhances accuracy by 0.55\%. In contrast, competing methods typically reduce accuracy by 0.42\%. These results confirm that FairTTTS effectively promotes more equitable decision-making while simultaneously improving predictive performance.
\end{abstract}

\keywords{Fairness in Machine Learning, Bias Mitigation, Tree Test Time Simulation, Monte Carlo, Post-processing Methods}

\maketitle
\section{Introduction}

The widespread adoption of machine learning (ML) algorithms in critical decision-making processes,  from hiring practices to judicial sentencing ~\cite{angwin2022machine} and credit scoring \cite{bolukbasi2016man}, has raised significant concerns about fairness and bias in automated systems. Despite their promise, ML models often reinforce biases in training data, leading to discriminatory outcomes for disadvantaged groups. For instance, predictive policing algorithms have been criticized for racial profiling, and automated loan approval systems have exhibited gender and racial disparities~\cite{angwin2016machine,hardt2016equality}. These issues underscore the urgent need for fairness-aware ML algorithms that can provide equitable outcomes without compromising predictive performance. Regulatory frameworks, such as the EU's General Data Protection Regulation (GDPR\footnote{https://eur-lex.europa.eu/eli/reg/2016/679/oj}) and guidelines from the U.S. Equal Employment Opportunity Commission (EEOC\footnote{https://www.eeoc.gov/prohibited-employment-policiespractices}), emphasize the importance of preventing algorithmic discrimination. Consequently, there is a growing demand for ML models that can make accurate predictions while adhering to fairness constraints, ensuring that protected groups are not disadvantaged by automated decisions.

Achieving fairness in production ML presents several challenges:

\begin{itemize}
    \item \textbf{Trade-off Between Accuracy and Fairness}: Improving fairness often leads to a decrease in predictive accuracy, as the model's ability to fit the data is constrained by fairness considerations~\cite{menon2018cost, friedler2019comparative}, specifically in post-processing bias mitigation methods.
    \item \textbf{Dependence on Sensitive Attributes}: Many fairness-enhancing methods require access to sensitive attributes during training or inference, which may not be feasible due to privacy concerns or legal restrictions~\cite{kamishima2012fairness}.
    \item \textbf{Metric Specificity}: Existing approaches are frequently tailored to specific fairness definitions (e.g., demographic parity, equalized odds), limiting their applicability across different contexts~\cite{hardt2016equality}.
    \item \textbf{Model and Data Constraints}: In practical settings, retraining models may not be possible due to proprietary models or limited access to the training data, necessitating non-invasive interventions that can adjust pre-trained models~\cite{pleiss2017fairness, zhang2018mitigating, barocas2019fairness}.

\end{itemize}

These challenges highlight the need for a flexible, generalizable approach that can enhance fairness without extensive retraining or specific metric dependencies.

\subsection{Our Solution: FairTTTS}

To address these challenges, we propose FairTTTS, a novel post-processing bias mitigation method designed to enhance fairness in classification tasks, particularly within decision tree models. Inspired by the TTTS method~\cite{Cohen_Arbili_Mirsky_Rokach_2024}, originally developed to improve robustness against adversarial attacks, FairTTTS adapts this approach to mitigate biases and promote equitable outcomes.

FairTTTS operates by introducing controlled randomness in the traversal paths of decision trees during inference. Specifically, it employs a Monte Carlo simulation-based technique that probabilistically flips decision paths at nodes involving protected attributes when unprivileged samples are directed towards unfavorable outcomes. This probabilistic adjustment is guided by a distance-based heuristic that considers the feature value's proximity to the decision threshold, ensuring that samples near the threshold have a higher chance of flipping, thus reducing potential biases encoded in the tree structure.


By decoupling fairness adjustments from the training process, FairTTTS offers these \textbf{contributions}:

\begin{itemize}

    \item \textbf{Fairness- and Accuracy-Enhancing Method}: We introduce FairTTTS, a post-processing bias mitigation method for ML classification tasks. Building upon the TTTS technique, which is designed to enhance accuracy, FairTTTS incorporates fairness considerations by probabilistically adjusting decision paths at protected attribute nodes. This dual focus enables FairTTTS to improve both fairness and accuracy, addressing a key limitation of many post-processing methods that typically involve trade-offs between these objectives.

    \item \textbf{Flexible Integration}: FairTTTS can be easily integrated into production ML systems, making it suitable for scenarios where retraining is impractical or when access to sensitive attributes is limited. This ensures compliance with privacy regulations and circumvents legal restrictions.

    \item \textbf{Broad Applicability}: Extensive experiments on seven benchmark datasets from different domains, demonstrate that FairTTTS achieves consistent improvements in fairness metrics (21\% reduction in Equalized Odds Difference) while maintaining or improving accuracy (average 0.55\% gain).
        
    \item \textbf{Metric-Agnostic Adaptability}: The approach is adaptable across various fairness metrics and constraints, providing the flexibility to address diverse application contexts without tailoring the solution to a single definition of fairness.

\end{itemize}

ion{Related Work} \label{sec:related_work}

\subsection{Bias Mitigation in Machine Learning}

Bias in machine learning can be mitigated by pre-processing, in-processing, and post-processing techniques\cite{barocas2019fairness}. Pre-processing techniques mitigate biases in training data by transforming protected attributes \cite{feldman2015certifying, calmon2017optimized, lum2016statistical}, learning fair representations\cite{samadi2018price}, or adjusting protected attribute distributions\cite{friedler2014certifying, cohen2024fairus}. In-processing methods incorporate fairness constraints directly into the model training process\cite{zafar2017fairness, abusitta2019generative} or impacting the model's architecture \cite{10.1145/3593013.3594008}. Post-processing methods, such as FairTTTS, adjust outputs from pre-trained models, providing flexibility when retraining is infeasible.

\subsubsection{Post-Processing Bias Mitigation Methods}


Post-processing methods address fairness after a model has been trained, making them particularly suitable for proprietary or pre-trained systems. Recent survey ~\cite{hort2024bias}, categorizes these methods into input correction, classifier correction, and output correction.

Input correction methods modify testing data before inference by applying perturbations on the test sample \cite{adler2018auditing}. Output Correction methods directly alter predicted labels. Reject option-based strategies \cite{kamiran2012decision} flip predictions near decision boundaries, prioritizing fairness for unprivileged groups. Confidence-based approaches \cite{fish2016confidence} use thresholds to modify predictions, often incorporating group-specific thresholds \cite{menon2018cost}. Techniques like causal model adjustments \cite{chiappa2019path} ensure fairness by aligning predictions with counterfactual worlds. Changing the label by understanding the Shapely values impact of the protected attribute \cite{harsh_kasyap__2024}. A recent study presented a method that balances predictions with synthetic data \cite{coheninger2025biasguardguardrailingfairnessmachine}. 

Classifier Correction methods adjust a trained model to improve fairness. Notable works include Hardt et al.'s Equalized Odds Post-Processing~\cite{hardt2016equality}, which optimizes predictions thresholds. Methods such as decision tree relabeling~\cite{kamiran2012decision} aim to balance fairness for decision trees.

Our work builds on classifier correction techniques, a Monte Carlo-based method that probabilistically adjusts decision paths during inference. Unlike previous methods relying on fixed thresholds or heuristics, FairTTTS dynamically incorporates fairness considerations at decision nodes, offering greater adaptability across diverse fairness metrics.

\subsection{Inspiration from TTTS Method}

The Tree Test Time Simulation (TTTS) method~\cite{Cohen_Arbili_Mirsky_Rokach_2024} introduces a probabilistic modification to decision paths in trees to enhance robustness against adversarial examples. TTTS employs Monte Carlo simulations to traverse alternative paths in the tree, guided by a distance-based heuristic that considers the feature value's proximity to the decision threshold. The flip probability is given by:

\begin{equation}
p_{\text{flip}} = \max\left(p_{\text{max\_threshold}} - \frac{|v - t|}{\sigma + \epsilon}, 0\right),
\end{equation}

where $v$ is the feature value, $t$ is the threshold at the node, $\sigma$ is the standard deviation of feature values at the node, and $p_{\text{max\_threshold}}$ is the maximum flip probability.

While TTTS focuses on robustness, we adapt its distance-based probabilistic traversal mechanism to address fairness concerns. By increasing the probability of flipping at nodes involving protected attributes when unprivileged groups are directed towards unfavorable outcomes, FairTTTS enhances fairness in decision-making. This adaptation allows for targeted adjustments that improve fairness metrics without compromising accuracy.

Our FairTTTS method introduces a more granular and adaptive mechanism. Instead of implementing fairness corrections at the final decision stage, FairTTTS exploits the internal structure of decision trees to provide more nuanced interventions and incorporates Monte Carlo simulations. Specifically, it leverages a threshold distance-based probabilistic traversal mechanism suggested by  TTTS for adversarial robustness and adjusts it for bias mitigation. In this mechanism, each node’s decision to flip the path is influenced by how close the feature value is to the node’s threshold, as well as by the presence of protected attributes directing unprivileged groups towards unfavorable outcomes. By integrating fairness considerations directly into the tree’s inference logic, FairTTTS can subtly adjust decision paths where biases are most pronounced and give more opportunities to the unprivileged group in each simulation, rather than applying a uniform threshold shift at the end of the pipeline.

\subsection{Measuring Bias in ML}

Fairness metrics are classified into two main categories\cite{caton2020fairness}: those based on outcome probabilities, such as Statistical Parity \cite{calders2010three} and Disparate Impact \cite{feldman2015certifying}, and those derived from the confusion matrix, including Equalized Odds, Equal Opportunity \cite{hardt2016equality} and Accuracy Rate Difference \cite{berk2021fairness}. A single scenario may be labeled fair by one metric and unfair by another \cite{verma2018fairness}. In this study, we focus on two widely used metrics, as identified in the survey~\cite{caton2020fairness}: Disparate Impact (DI) from the outcome-based category and Equalized Odds Difference (EOD) from the confusion-matrix-based category.



\section{Method} \label{sec:methods}

\subsection{Problem Formulation}

Let $\mathcal{D} = \{(X_i, y_i)\}_{i=1}^N$ be a dataset where $X_i \in \mathbb{R}^d$ represents the feature vector of the $i$-th instance, and $y_i \in \{0, 1\}$ is the corresponding binary label. A sensitive attribute $Z \subseteq X$ divides the dataset into privileged ($Z=1$) and unprivileged ($Z=0$) groups.

Our objective is to enhance both predictive accuracy and fairness for a given pre-trained classifier $\mathcal{M}: \mathbb{R}^d \rightarrow \{0, 1\}$. This involves improving the model's ability to make accurate predictions while reducing disparities in outcomes between privileged and unprivileged groups. Specifically, we aim to:

1. Maximize predictive accuracy, denoted as $\mathcal{A}(\mathcal{M})$, which measures the proportion of correct predictions made by $\mathcal{M}$. 

2. Simultaneously enhance fairness, denoted as $\mathcal{F}(\mathcal{M})$, where fairness metrics such as disparate impact and equalized odds difference quantify the model's adherence to equity principles.

Formally, our goal is to adjust the decision paths during inference to achieve maximum $\mathcal{A}(\mathcal{M}) \quad \text{and} \quad$ maximum $\mathcal{F}(\mathcal{M})$.

where $\mathcal{F}(\mathcal{M})$ measures fairness improvements beyond satisfying minimum fairness constraints. This dual objective ensures that the method achieves a balance between accuracy and fairness, surpassing traditional approaches that prioritize accuracy while meeting fairness constraints as secondary objectives.

The proposed FairTTTS framework operates in a post-processing manner, making it applicable to various pre-trained models without requiring retraining. By probabilistically adjusting decision paths based on a distance-based heuristic and fairness considerations, FairTTTS ensures more equitable outcomes while maintaining or improving predictive performance.

\subsection{Overview of FairTTTS}

\textbf{FairTTTS} adapts TTTS to address fairness by introducing controlled randomness during the decision-making process, specifically targeting nodes involving protected attributes.

The key idea is to increase the probability of flipping the traversal direction at these nodes when an unprivileged sample is being directed toward an unfavorable outcome. This probabilistic adjustment aims to mitigate bias encoded in the tree structure, providing unprivileged groups with increased opportunities for favorable classification without retraining the model.

\begin{figure}[ht!]
    \centering
    \includegraphics{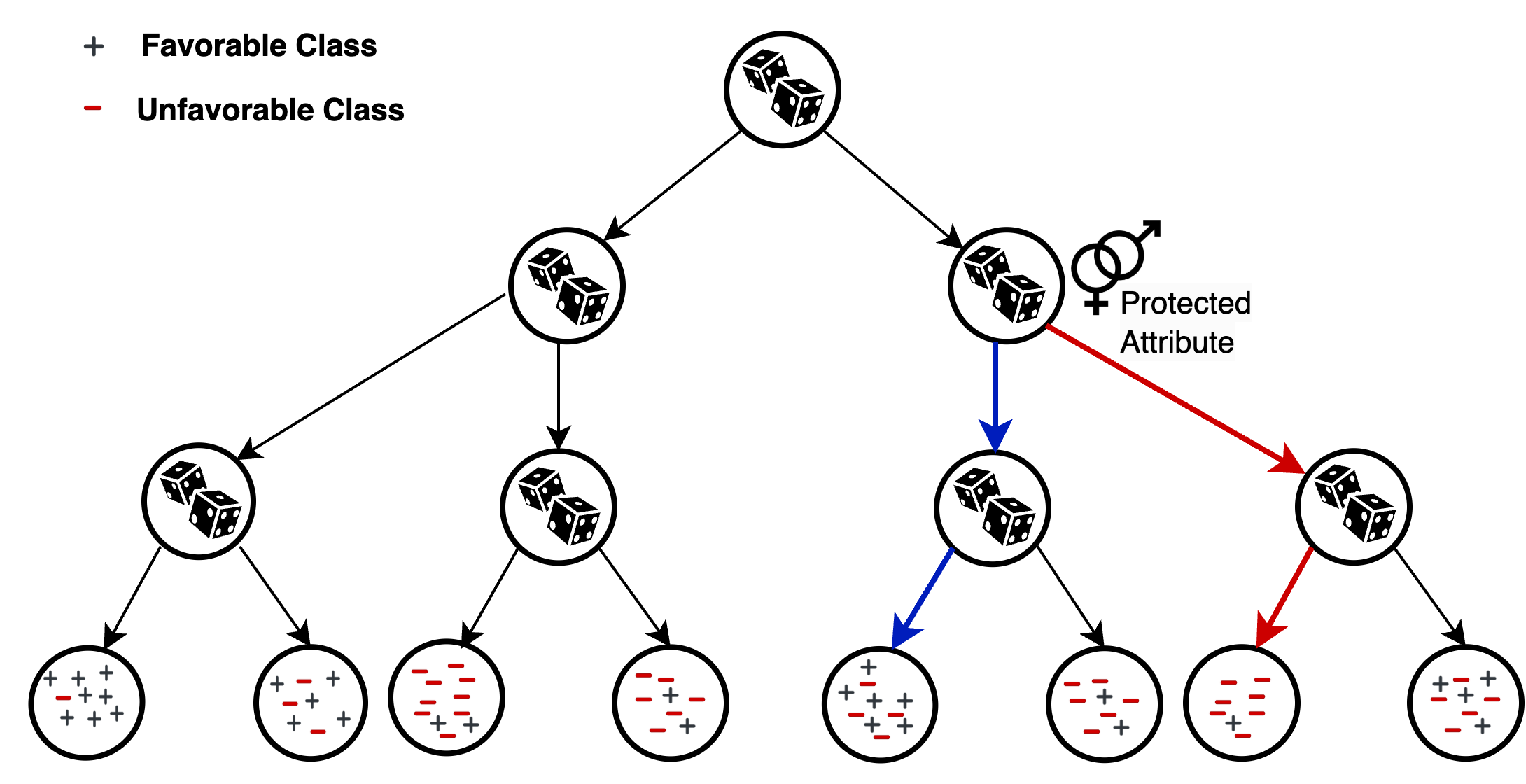}
    \caption{An illustrative example of the FairTTTS approach. At nodes involving protected attributes (shown at the second depth right node), unprivileged samples are directed toward the unfavorable class. By increasing the flipping probability by a factor \(\alpha\) at these nodes, FairTTTS probabilistically redirects more of these samples toward favorable leaf nodes, effectively providing greater opportunities to the unprivileged group.}
    \label{fig:fairttts_illustration}
\end{figure}

As illustrated in Figure~\ref{fig:fairttts_illustration}, when the decision tree encounters a protected attribute node that directs unprivileged samples towards an unfavorable class, FairTTTS increases the flipping probability by a factor \(\alpha\). This adjustment ensures that a higher proportion of these samples have the chance to be redirected to a more favorable class. By doing so, the algorithm dynamically reduces biases that may have been encoded in the original tree structure, allowing unprivileged groups more equitable access to favorable outcomes without retraining the underlying model.

\subsection{Formal Description of the Method}

\subsubsection{Decision Tree Structure}

Consider a decision tree $\mathcal{T}$ trained on $\mathcal{D}$ using standard algorithms (e.g., CART). Each non-leaf node $n$ in $\mathcal{T}$ splits on a feature $f_n$ with a threshold $t_n$, directing samples to its left or right child nodes based on the feature value.

\subsubsection{Monte Carlo Simulations for Fairness}

For a given test sample $X$, we perform $S$ Monte Carlo simulations to traverse $\mathcal{T}$. In each simulation, we probabilistically decide whether to follow the original traversal path or to flip to the opposite child at each node, based on a calculated flip probability $p_{\text{flip}}$.

\paragraph{Flip Probability Calculation}
At each node $n$, we compute the flip probability $p_{\text{flip}}(n, X)$ using a distance-based heuristic:
\begin{equation}
p_{\text{flip}}(n, X) = \min\left(p_{\max} - \frac{|X_{f_n} - t_n|}{\delta_{\max}}, \ p_{\max}\right),
\end{equation}
where:
\begin{itemize}
    \item $X_{f_n}$ is the value of feature $f_n$ in sample $X$.
    \item $\delta_{\max}$ is the maximum possible distance between feature values and thresholds in the tree, ensuring $p_{\text{flip}} \geq 0$.
    \item $p_{\max}$ is the maximum flip probability, set as a hyperparameter (e.g., $0.1$).
\end{itemize}

\paragraph{Fairness Adjustment at Protected Nodes}

If node $n$ splits on the protected attribute $Z$, and the sample $X$ belongs to the unprivileged group ($X_Z = 0$), we adjust the flip probability when the traversal is directing $X$ towards the unfavorable class $y = 0$. Specifically, we increase $p_{\text{flip}}$ by a factor $\alpha > 1$:

\begin{equation}
p_{\text{flip}}(n, X) = \min(\alpha \cdot p_{\text{flip}}(n, X), \ 0.5).
\end{equation}

This adjustment ensures that unprivileged samples have a higher chance of being redirected towards a favorable outcome at critical decision points.

\subsubsection{Traversal Algorithm}

During each simulation, we traverse the tree from the root node according to Algorithm~\ref{alg:traverse}.

\begin{algorithm}[H]
\caption{Probabilistic Tree Traversal for Sample $X$}
\label{alg:traverse}
\begin{algorithmic}[1]
\Procedure{Traverse}{$n$, $X$}
    \If{$n$ is a leaf node}
        \State \Return predicted class at $n$
    \Else
        \State Compute $p_{\text{flip}}(n, X)$
        \State Decide whether to flip traversal with probability $p_{\text{flip}}(n, X)$
        \If{not flipping}
            \If{$X_{f_n} \leq t_n$}
                \State $n_{\text{next}} \leftarrow$ left child of $n$
            \Else
                \State $n_{\text{next}} \leftarrow$ right child of $n$
            \EndIf
        \Else
            \If{$X_{f_n} \leq t_n$}
                \State $n_{\text{next}} \leftarrow$ right child of $n$
            \Else
                \State $n_{\text{next}} \leftarrow$ left child of $n$
            \EndIf
        \EndIf
        \State \Return \Call{Traverse}{$n_{\text{next}}$, $X$}
    \EndIf
\EndProcedure
\end{algorithmic}
\end{algorithm}

\subsubsection{Aggregation of Predictions}

After $S$ simulations, we obtain a set of predicted classes $\{\hat{y}_1, \hat{y}_2, \dots, \hat{y}_S\}$ for sample $X$. The final predicted probability for class $c$ is computed as:

\begin{equation}
P(\hat{y} = c \mid X) = \frac{1}{S} \sum_{s=1}^{S} \mathbb{I}[\hat{y}_s = c],
\end{equation}

where $\mathbb{I}[\cdot]$ is the indicator function.

\subsection{Rationale Behind the Method}

\subsubsection{Why We Use Probabilistic Traversal}

Decision trees can encode biases present in the training data, particularly when splitting on protected attributes. By introducing randomness in the traversal paths, we can disrupt patterns that lead to unfair treatment of unprivileged groups.

\subsubsection{Role of Distance-Based Heuristic}

The distance-based heuristic ensures that the probability of flipping is higher when the feature value is close to the decision threshold, indicating uncertainty in the decision. This aligns with the intuition that near-threshold decisions are more susceptible to bias.

\subsubsection{Fairness Enhancement through $\alpha$ Adjustment}

Increasing the flip probability by a factor $\alpha$ at protected attribute nodes for unprivileged samples directed towards unfavorable outcomes actively counteracts the bias. It provides these samples with additional opportunities to reach favorable leaf nodes, improving fairness metrics.


\subsection{Theoretical Considerations and Fairness Guarantees} 

Although decision trees are discrete structures and flipping decisions introduces stochasticity, we can provide intuition and partial theoretical grounding for why FairTTTS improves fairness. Consider a binary classification setting with a sensitive attribute $Z$. Suppose the original decision tree is fixed and let $\hat{Y}$ be the random variable denoting the predicted label of a sample $X$. Without loss of generality, assume higher flips at protected nodes reduce the expected rate at which unprivileged samples are directed to $Y=0$ leaves.

\textbf{Key Idea:} The probability of flipping a decision at a protected node effectively shifts the local decision boundary around that node. By selectively increasing flip probabilities for unprivileged samples near the decision threshold, we create a locally “more inclusive” region where unprivileged individuals have a higher probability of receiving favorable outcomes. This translates to a reduction in group-level disparity.

\textbf{Example:}  
Consider a single protected-attribute node with threshold $t$ splitting an unprivileged group $U$. Let $\hat{Y}$ be the prediction without flipping and $\hat{Y}^\prime$ the prediction after flipping. If $X_{f_n}$ is uniformly distributed near $t$ for samples in $U$, increasing $p_{\text{flip}}$ by $\alpha$ increases the probability that a sample close to $t$ will be assigned to the branch leading to $Y=1$ leaves. Hence, for these samples, 
\[
P(\hat{Y}^\prime=1 \mid Z=0) \geq P(\hat{Y}=1 \mid Z=0) + \Delta,
\]
for some $\Delta>0$ dependent on $\alpha$ and the distribution of $X_{f_n}$ near $t$. When combined over multiple nodes, these local shifts can globally improve fairness metrics like disparate impact or reduce the difference in TPR/FPR between groups.

While this argument is heuristic, it suggests that under mild assumptions (e.g., local continuity of feature distributions, reasonable threshold assignments), increasing flip probabilities at protected nodes can raise the likelihood of favorable outcomes for the unprivileged group. In turn, this reduces disparities at the aggregate level.

\section{Experimental Setup} \label{sec:experimental_setup}

\subsection{Datasets}

We evaluate FairTTTS on seven distinct datasets commonly used in fairness research across eight total experiments (splitting the Adult dataset by race and sex) as covered in Table \ref{tab:datasets}:

\begin{table}[ht]
    \centering
    \caption{Description of Datasets Used for Experiments}
    \label{tab:datasets}
    \begin{tabular}{p{2.6cm}p{1cm}p{1.2cm}p{4cm}p{3.3cm}}
        \hline
        \textbf{Dataset} & \textbf{Samples} & \textbf{Attributes} & \textbf{Description} & \textbf{Protected Attribute(s)} \\
        \hline
        Adult Income~\cite{adult_2} & 32561 & 15 & Predicts whether an individual's income exceeds \$50K/year. & \texttt{race} (ADULT\_RACE),  \texttt{sex} (ADULT\_SEX) \\
        \hline
        Bank Marketing~\cite{moro2014data} & 11162 & 17 & Predicts whether a client will subscribe to a term deposit. & \texttt{age} (BANK\_AGE) \\
        \hline
        COMPAS~\cite{larson2016we} & 7217 & 15 & Predicts recidivism risk scores used in criminal justice. & \texttt{race} (COMPAS\_RACE) \\
        \hline
        German Credit~\cite{kohavi1996scaling} & 1000 & 10 & Assesses credit risk. & \texttt{sex} (CREDIT\_SEX) \\
        \hline
        Diabetes\footnote{Medical dataset that was published in 2023 on Kaggle. The data set is created by Mohammed Mustafa.} & 100000 & 9 & Predicts hospital readmission for diabetic patients. & \texttt{age} (DIABETES\_AGE) \\
        \hline
        MIMIC~\cite{syed2021application} & 1176 & 51 & Predicts Intensive-Care Unit patients' mortality. & \texttt{sex} (MIMIC\_SEX) \\
        \hline
        Recruiting\footnote{Students' admission to university dataset that was published in 2022 on Kaggle. The data set is created by Sieuwert van Otterloo.} & 4000 & 15 & Simulated dataset for recruitment decisions. & \texttt{sex} (RECRUIT\_SEX) \\
        \hline
    \end{tabular}
\end{table}

\subsection{Evaluation Metrics}

To assess the performance of FairTTTS, we use the following metrics:

\begin{itemize}
    \item \textbf{Equalized Odds Difference (EOD)}: 
    Equalized Odds aims to ensure that both privileged (priv) and unprivileged (unpriv) groups have similar true positive rates (TPR) and false positive rates (FPR). Let $\text{TPR}_{priv}$, $\text{TPR}_{unpriv}$, $\text{FPR}_{priv}$, and $\text{FPR}_{unpriv}$ denote these rates for the privileged and unprivileged groups, respectively. We define EOD as:
    \begin{equation}
    \text{EOD} = \frac{\left| \text{TPR}_{unpriv} - \text{TPR}_{priv} \right| + \left| \text{FPR}_{unpriv} - \text{FPR}_{priv} \right|}{2}
    \end{equation}
    Lower EOD values indicate better fairness.

    \item \textbf{Disparate Impact (DI)}: 
    Disparate Impact measures the ratio of favorable outcome probabilities between unprivileged and privileged groups. Let $P(Y=1 \mid unpriv)$ and $P(Y=1 \mid priv)$ denote the probabilities of a positive (favorable) outcome for the unprivileged and privileged groups, respectively. Then:
    \begin{equation}
    \text{DI} = \frac{P(Y=1 \mid unpriv)}{P(Y=1 \mid priv)}
    \end{equation}
    A DI close to $1$ indicates fairness, while values significantly below $1$ indicate potential bias.

   \item \textbf{Accuracy}: 
    Accuracy is the proportion of correctly classified instances out of all instances. Let $TP$ be the number of true positives, $TN$ the number of true negatives, $FP$ the number of false positives, and $FN$ the number of false negatives. Then:
    \begin{equation}
    \text{Accuracy} = \frac{TP + TN}{TP + TN + FP + FN}.
    \end{equation}
    Higher accuracy values indicate better predictive performance.
\end{itemize}

\subsection{Comparison Methods}

We compare FairTTTS with the following methods:

\begin{itemize}
    \item \textbf{Baseline}: Standard random forest classifier without fairness interventions.
    \item \textbf{ThresholdOptimizer}~\cite{hardt2016equality}: Post processing method that adjusts decision thresholds of the protected attribute to improve fairness.

\end{itemize}

\subsection{Model}

For the evaluation of FairTTTS, we tested the method on both Random Forest and standalone Decision Tree classifiers. Random Forests are widely used due to their strong predictive performance and interpretability, making them an ideal candidate for demonstrating the effectiveness of our approach. Similarly, Decision Trees offer a simpler yet highly interpretable architecture, further showcasing the versatility of FairTTTS across different decision tree-based models. Since FairTTTS operates directly on the decision tree structure, its application is straightforward: the method identifies splits involving protected attributes and probabilistically adjusts decision paths during inference. This ensures improved fairness without retraining or modifying the underlying model architecture. Importantly, FairTTTS is not restricted to these models. It is applicable to any decision tree-based architecture, including Gradient Boosted Trees and frameworks like LightGBM and XGBoost.

\subsection{Implementation Details}

\subsubsection{Hyperparameters}

For FairTTTS, we set: Number of simulations $S = 100$. Maximum flip probability $p_{\max} = 0.1$. Fairness adjustment factor $\alpha = 9.0$. These hyperparameters were chosen based on preliminary experiments to balance accuracy and fairness.

\subsubsection{Additional Parameters}

For each dataset, we define:
\begin{itemize}
     \item Protected Attribute: As specified in the dataset description.
     \item Privileged Group: Samples where the protected attribute equals $1$.
     \item Unprivileged Group: Samples where the protected attribute equals $0$.
     \item Favorable Class: The positive class label ($y = 1$).
     \item Unfavorable Class: The negative class label ($y = 0$).
\end{itemize}



\subsection{Experimental Process}

\begin{enumerate}
    \item Data Preprocessing: We handle missing values, encode categorical variables, and split each dataset into 5 folds cross-validation.
    \item Model Training: Train the Random\_Forest\_Baseline on the training set.
    \item Fairness Enhancements:
        \begin{itemize}
            \item Apply ThresholdOptimizer to the baseline model.
            \item Implement FairTTTS by wrapping the trained decision tree with our Monte Carlo traversal method.
        \end{itemize}
    \item Evaluation: Predict on the test set using each method and compute the evaluation metrics and variances.
\end{enumerate}

\section{Results} \label{sec:results}

We evaluated FairTTTS against the Baseline (RandomForest) and ThresholdOptimizer across eight experiments. These experiments were conducted on datasets with various sensitive attributes (e.g., race, sex, age): ADULT\_RACE, ADULT\_SEX, BANK\_AGE, COMPAS\_RACE, CREDIT\_SEX, DIABETES\_AGE, MIMIC\_SEX, and RECRUIT\_SEX. Our primary fairness metrics were Equalized Odds Difference (EOD) and Disparate Impact (DI), followed by Accuracy as a secondary metric.

Table~\ref{tab:results_highlighted} displays the mean and standard deviation (Mean ± STD) for EOD, DI, and Accuracy across all experiments. For each experiment, the best-performing method in terms of EOD, DI, or Accuracy is shown in \textbf{bold}. Additionally, each cell associated with FairTTTS is color-coded based on its performance relative to both the baseline and ThresholdOptimizer: {\color[HTML]{008000}Green} cell: FairTTTS outperforms both the baseline and ThresholdOptimizer on that metric. 

\subsection{Key Observations}

\textbf{Fairness improvements}, measured primarily by EOD and supported by DI, was our main focus. Across all eight experiments FairTTTS reduced EOD compared to the baseline in all eight cases. In seven of these, FairTTTS also outperformed ThresholdOptimizer. On average, FairTTTS improved fairness (indicated by reduced EOD) by 21\% over the baseline, while ThresholdOptimizer improved it only by 18.77\%.  FairTTTS improved DI over the baseline in seven out of eight experiments. In three of these, it also surpassed the ThresholdOptimizer. 

EOD reductions confirm that FairTTTS is providing more equitable decision-making by ensuring that privileged and unprivileged groups face more similar error rates. DI improvements signal a more balanced distribution of positive outcomes. Together, these fairness metrics, underscore that FairTTTS is meeting its primary objective of enhancing fairness in decision-making.

\textbf{Accuracy improvements} were also observed: FairTTTS increased Accuracy over both the baseline and ThresholdOptimizer in five of the eight experiments, on average improving Accuracy by  0.55\% compared to the baseline. In contrast, ThresholdOptimizer often reduced Accuracy by 0.42\%. 

\textbf{Running Time Results:} We evaluated inference speed for the Random\_Forest\_Baseline, ThresholdOptimizer, and FairTTTS methods across eight tabular datasets and protected attributes. For the baseline and ThresholdOptimizer methods, total inference time across the entire dataset was typically around 0.2 seconds for the Adult dataset and even less for smaller datasets. FairTTTS, however, required around 52 seconds in total on Adult—roughly 200 to 250 times longer than the other methods. Although this may appear substantial at first glance, it is important to note that the datasets are tabular, which generally means that even these extended running times remain small per individual sample. For instance, the 52-second total on the Adult dataset translates to about 1.6 milliseconds per inference. In most real-world scenarios where tabular data processing does not require extensive computational resources, this overhead is still negligible. For use cases where fairness is a critical requirement, the slight latency increase per sample may be a reasonable trade-off.

\subsection{Summary of Results}

In summary,Table \ref{tab:results_highlighted} clearly shows that FairTTTS delivers substantial improvements in fairness over both the baseline and ThresholdOptimizer in most experiments. It achieves these gains while often maintaining or increasing Accuracy. The consistent patterns confirm that FairTTTS is a practical, effective post-processing technique for achieving more equitable decision-making in decision tree models.

\begin{table*}[ht]
\centering
\caption{Performance comparison of methods across eight experiments using Random Forest. Bolded values indicate the best results for accuracy and equalized odds difference within each experiment. Green items indicate FairTTTS superiority.}
\label{tab:results_highlighted}
\resizebox{\textwidth}{!}{%
\begin{tabular}{@{}llrrrrrr@{}}
\toprule
Dataset\_Attribute              & Method                 & \multicolumn{1}{l}{Accuracy}  & \multicolumn{1}{l}{Accuracy} & \multicolumn{1}{l}{Equalized Odds} & \multicolumn{1}{l}{Equalized Odds} & \multicolumn{1}{l}{Disparate Impact} & \multicolumn{1}{l}{Disparate Impact} \\ 
                                &                        & \multicolumn{1}{l}{mean}      & \multicolumn{1}{l}{std}      & \multicolumn{1}{l}{mean}           & \multicolumn{1}{l}{std}            & \multicolumn{1}{l}{mean}             & \multicolumn{1}{l}{std}              \\ \midrule
                                & Baseline & 0.8581                        & 0.0025                       & 0.0707                             & 0.0598                             & 0.4374                               & 0.0846                               \\
                                & ThresholdOptimizer     & 0.8581                        & 0.0029                       & 0.0663                             & 0.0381                             & \textbf{0.4632}                      & 0.0739                               \\ 
\multirow{-3}{*}{ADULT\_RACE}   & FairTTTS & {\color[HTML]{008000}\textbf{0.8596}}               & 0.0048                       & {\color[HTML]{008000}\textbf{0.0656}}                    & 0.0370                             &  0.4568       & 0.0733                               \\ \hline
                                & Baseline & 0.8593                        & 0.0009                       & 0.0787                             & 0.0405                             & 0.3115                               & 0.0302                               \\
                                & ThresholdOptimizer     & 0.8586                        & 0.0009                       & 0.0681                             & 0.0396                             & \textbf{0.3491}                      & 0.0619                               \\
\multirow{-3}{*}{ADULT\_SEX}    & FairTTTS & {\color[HTML]{008000}\textbf{0.8594}}               & 0.0037                       & {\color[HTML]{008000}\textbf{0.0666} }                   & 0.0400                             &  0.3251        & 0.0383                               \\ \hline
                                & Baseline & \textbf{0.8532}                        & 0.0064                       & 0.0275                             & 0.0218                             & 1.0185                               & 0.0430                               \\
                                & ThresholdOptimizer     & 0.8526                        & 0.0060                       & 0.0225                             & 0.0216                             & 1.0342                               & 0.0409                               \\
\multirow{-3}{*}{BANK\_AGE}     & FairTTTS &  0.8439 & 0.0029                       & {\color[HTML]{008000}\textbf{0.0103} }                   & 0.0166                             & {\color[HTML]{008000}\textbf{1.0515} }                     & 0.0349                               \\ \hline
                                & Baseline & 0.6325                        & 0.0112                       & 0.2028                             & 0.0399                             & 1.5949                               & 0.0458                               \\
                                & ThresholdOptimizer     & 0.6167                        & 0.0159                       & \textbf{0.0505}                    & 0.0668                             & 1.1957                               & 0.0681                               \\
\multirow{-3}{*}{COMPAS\_RACE}  & FairTTTS & {\color[HTML]{008000}\textbf{0.6652} }              & 0.0129                       & { 0.1865}      & 0.0410                             & {\color[HTML]{008000}\textbf{1.6405} }                     & 0.1195                               \\ \hline
                                & Baseline & \textbf{0.6665}                        & 0.0636                       & 0.0993                             & 0.1047                             & 0.8638                               & 0.1841                               \\
                                & ThresholdOptimizer     & 0.6646                        & 0.0631                       & 0.1220                             & 0.1055                             & 0.8416                               & 0.1272                               \\
\multirow{-3}{*}{CREDIT\_SEX}   & FairTTTS &  0.6589 & 0.0384                       & {\color[HTML]{008000}\textbf{0.0911}}                    & 0.0890                             & {\color[HTML]{008000}\textbf{0.9404} }                     & 0.1555                               \\ \hline
                                & Baseline & 0.9675                        & 0.0014                       & 0.0479                             & 0.0402                             & 0.0946                               & 0.0100                               \\
                                & ThresholdOptimizer     & 0.9638                        & 0.0014                       & 0.0391                             & 0.0364                             & \textbf{0.1392}                      & 0.0136                               \\
\multirow{-3}{*}{DIABETES\_AGE} & FairTTTS & {\color[HTML]{008000}\textbf{0.9718} }              & 0.0005                       & {\color[HTML]{008000}\textbf{0.0346}}                    & 0.0357                             & { 0.1032}        & 0.0117                               \\ \hline
                                & Baseline & 0.8776                        & 0.0090                       & 0.1526                             & 0.1141                             & 0.5236                               & 0.6543                               \\
                                & ThresholdOptimizer     & 0.8793                        & 0.0076                       & 0.1268                             & 0.0971                             & \textbf{1.0014}                      & 0.4758                               \\ 
\multirow{-3}{*}{MIMIC\_SEX}    & FairTTTS & {\color[HTML]{008000}\textbf{0.8801} }              & 0.0077                       & {\color[HTML]{008000}\textbf{0.1259} }                   & 0.0680                             & { 0.9479}        & 0.8441                               \\ \hline
                                & Baseline & \textbf{0.8420}               & 0.0074                       & 0.0941                             & 0.0453                             & 0.7072                               & 0.0614                               \\
                                & ThresholdOptimizer     & 0.8400                        & 0.0083                       & 0.0731                             & 0.0389                             & \textbf{0.7514}                      & 0.0311                               \\
\multirow{-3}{*}{RECRUIT\_SEX}  & FairTTTS &  0.8320 & 0.0111                       & {\color[HTML]{008000}\textbf{0.0570} }                   & 0.0425                             &  0.6935        & 0.0409                               \\ \midrule 
\end{tabular}}
\end{table*}

\subsubsection{Consistency and Robustness}

FairTTTS demonstrates consistent improvements in both fairness and accuracy across diverse datasets, sensitive attributes (race, sex, age), and varying data conditions such as class imbalance. It frequently outperforms the baseline and ThresholdOptimizer, and remains competitive even when not surpassing the latter. Overall, FairTTTS reduces bias (lower EOD), improves outcome balance (higher DI), and often boosts accuracy, confirming its versatility and robustness.

\subsubsection{Comparative Analysis with Related Work}

Most post-processing bias mitigation methods, such as ThresholdOptimizer, improve fairness metrics but often at the cost of accuracy. In our experiments, the ThresholdOptimizer method improved fairness metrics in some cases but resulted in an average accuracy drop of 0.42\%. In contrast, FairTTTS achieved superior fairness in seven out of eight experiments while also maintaining or improving accuracy. In the one experiment where ThresholdOptimizer outperformed FairTTTS on the fairness metric, FairTTTS delivered better accuracy.

\begin{figure}[ht]
    \centering
    \includegraphics[width=\textwidth]{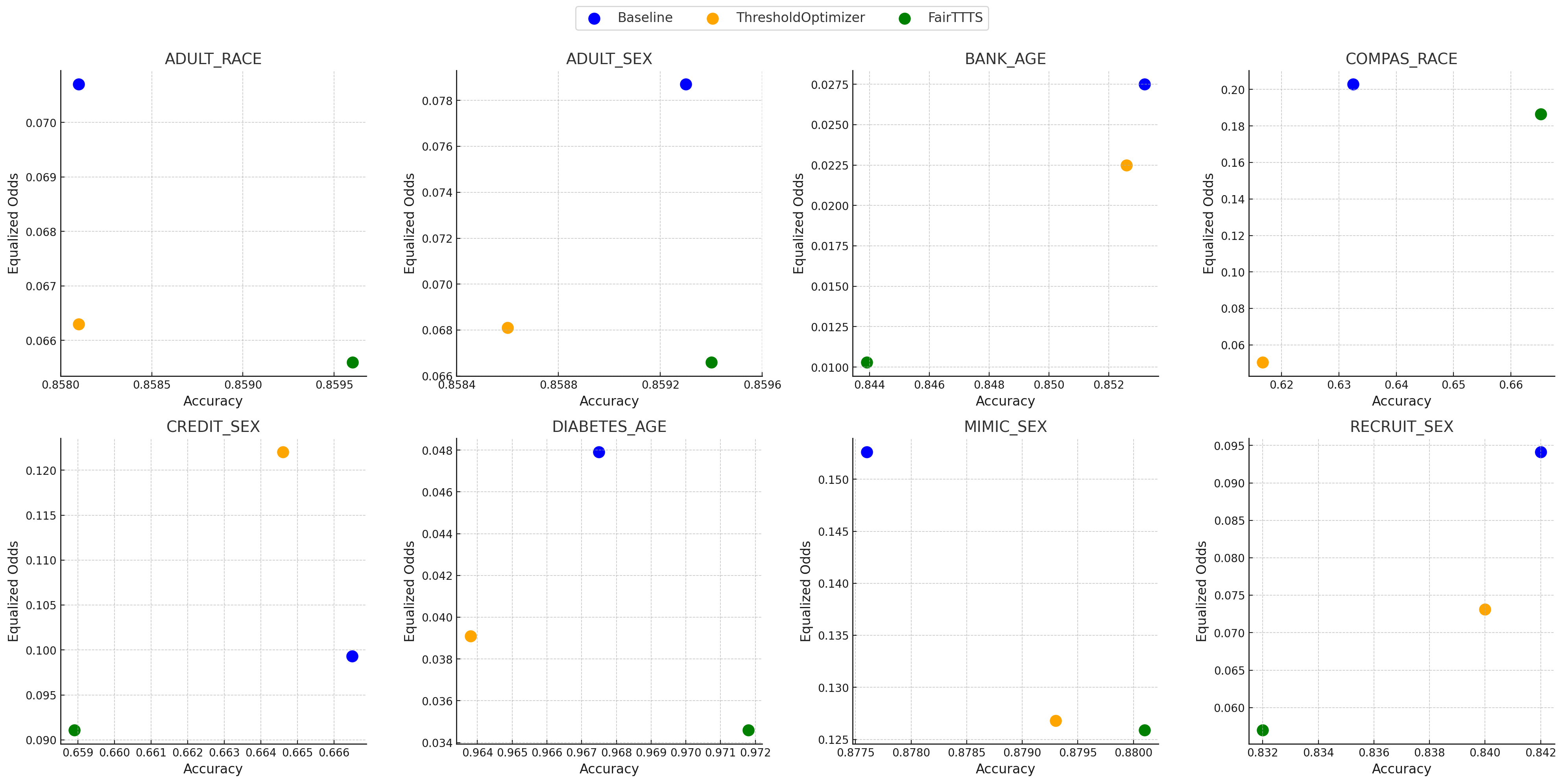}
    \caption{Accuracy  vs. Equalized Odds Across Datasets, Protected Attributes and Methods for \textbf{Random Forest} Architecture. Each plot represents a particular experiment, showing the performance of different methods in different colors. Notably, in seven out of eight experiments, FairTTTS achieves the best fairness, and in all eight experiments, it performs better than the baseline in terms of fairness outcomes.}

    \label{fig:accuracy_vs_eod}
\end{figure}

Figure~\ref{fig:accuracy_vs_eod} illustrates the trade-offs between accuracy and Equalized Odds across experiments (combinations of datasets and protected attributes) for Random Forest. Methods are distinguished by color. Unlike traditional methods that often require sacrificing accuracy for fairness, FairTTTS leverages its foundation in the TTTS technique, which is inherently designed to optimize accuracy. As a result, FairTTTS effectively balances fairness and accuracy across a variety of datasets, demonstrating its robustness and generalizability in diverse scenarios.

\begin{figure}[ht]
    \centering
    \includegraphics[width=\textwidth]{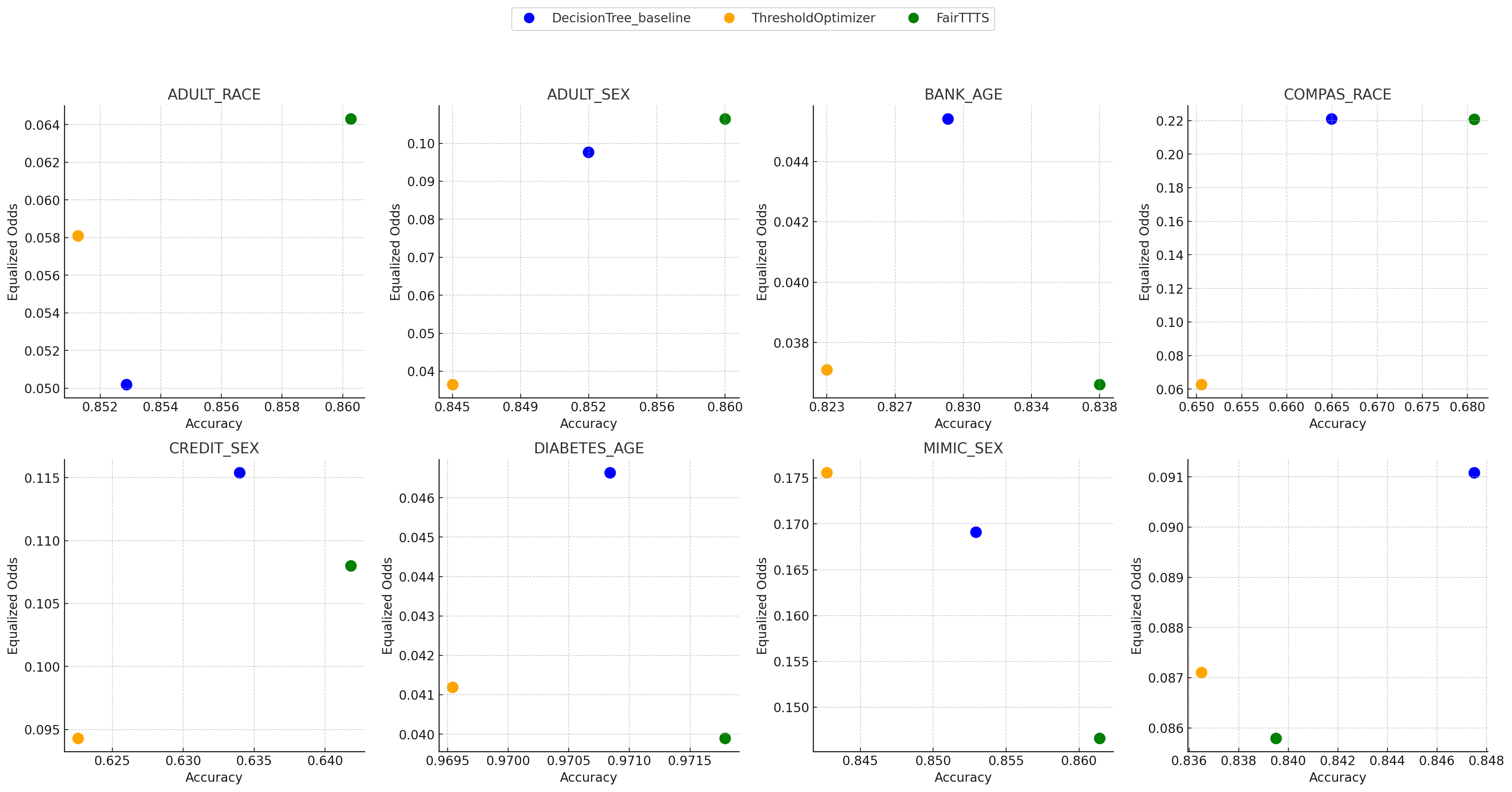}
    \caption{Accuracy vs. Equalized Odds Difference across datasets, protected attributes, and methods for \textbf{Decision Tree} architectures. Each plot represents a particular experiment, showing the performance of different methods in distinct colors. Notably, in seven out of eight experiments, FairTTTS achieves the highest accuracy. In six out of eight experiments, fairness improves compared to the baseline, and in four out of eight experiments, fairness improves compared to related work.}
    \label{fig:accuracy_vs_eod_DT}
\end{figure}

FairTTTS is applicable to any decision tree model architecture. The experimental results demonstrate the method's superiority in mitigating bias while also improving accuracy, particularly when applied to Decision Tree models, as illustrated in Figure~\ref{fig:accuracy_vs_eod_DT}.

\subsection{Sensitivity Analysis}

\begin{figure}[ht]
    \centering
    \includegraphics[width=0.98\textwidth]{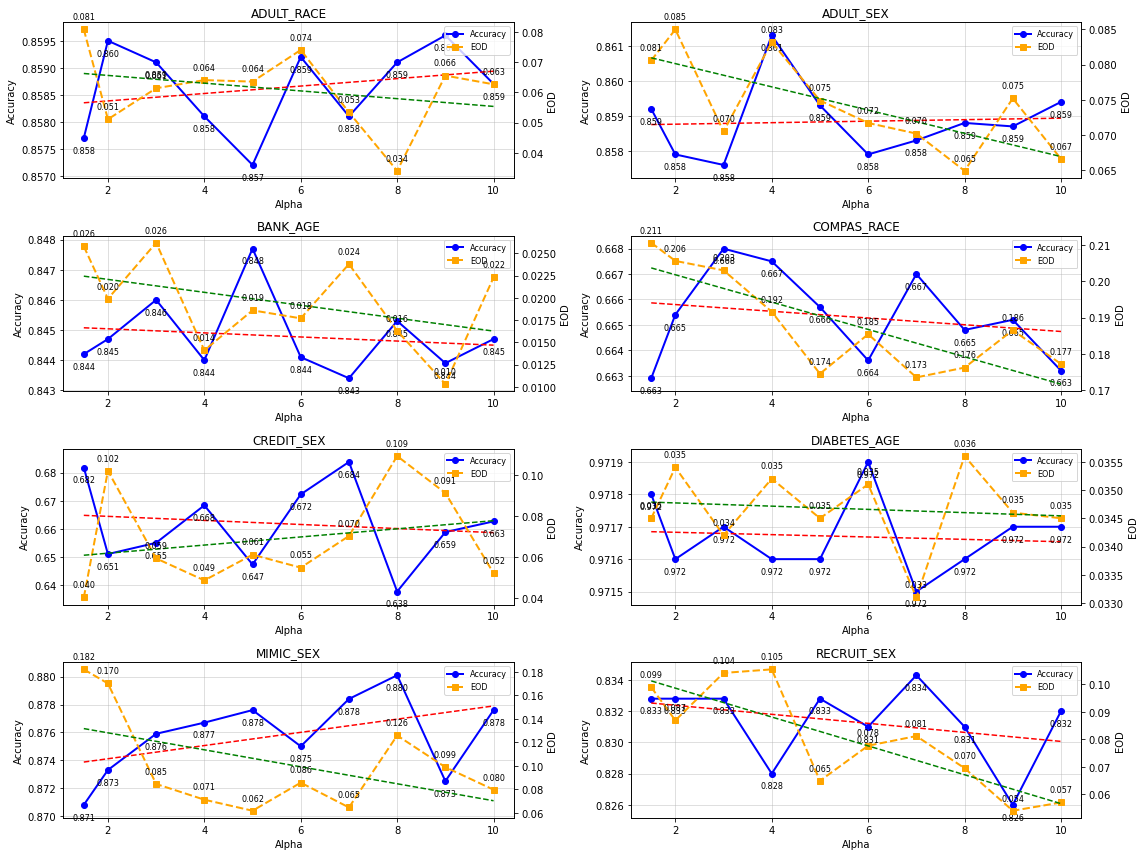}
    \caption{Sensitivity of FairTTTS to the Fairness Adjustment Factor \(\alpha\): The figure illustrates the effects of varying \(\alpha\) on Accuracy (solid blue lines, left vertical axis) and Equalized Odds Difference (dashed orange lines, right vertical axis). Trend lines (red for Accuracy and green for EOD) highlight the general patterns, showing that moderate \(\alpha\) values often achieve improved fairness (lower EOD) without significantly compromising Accuracy, as indicated by the downward slopes of the trend lines.}
 
    \label{fig:alpha_sensitivity}
\end{figure}

To evaluate the impact of the fairness adjustment factor \(\alpha\) on FairTTTS, we performed a detailed sensitivity analysis across all datasets, presented in Figure \ref{fig:alpha_sensitivity}. The parameter \(\alpha\) controls the probability of flipping decisions at nodes associated with protected attributes, influencing both fairness and accuracy. By varying \(\alpha\) from low to high values, we examined its effects comprehensively.

The results indicate that FairTTTS generally maintains stable accuracy performance across different \(\alpha\) settings. Moderate \(\alpha\) values consistently provide the best trade-off between fairness and accuracy, achieving significant reductions in Equalized Odds Difference without adversely affecting accuracy. At lower \(\alpha\) values, the model exhibits modest but meaningful fairness improvements over the baseline. As \(\alpha\) increases, disparities between groups diminish more significantly, leading to enhanced fairness as evidenced by lower EOD values. Importantly, these gains in fairness typically do not come at a steep cost to accuracy; in many cases, accuracy remains stable or improves slightly at moderate \(\alpha\) levels.

However, pushing \(\alpha\) to extremely high values does not necessarily lead to continuous improvements. Overly aggressive flipping probabilities can introduce too much randomness. This suggests that, while FairTTTS offers a strong mechanism to enhance fairness, excessively large \(\alpha\) values may lead to diminishing returns.

Our analysis further reveals that the optimal \(\alpha\) value depends on the dataset. While some datasets reach a favorable trade-off at lower \(\alpha\) settings, others require fine-tuning to achieve similar outcomes. Practitioners are advised to begin with moderate \(\alpha\) values and adjust based on the specific fairness requirements and accuracy thresholds of their application domain.

In summary, the sensitivity analysis demonstrates that FairTTTS offers substantial flexibility in balancing fairness and accuracy through the tuning of \(\alpha\). By appropriately calibrating \(\alpha\), users can achieve significant fairness enhancements with minimal or no loss of accuracy, establishing FairTTTS as a versatile tool for fairness-aware machine learning across diverse real-world scenarios.

\section{Conclusion} \label{sec:conclusion}

\subsection{Limitations and Future Work}

While FairTTTS is effective for enhancing fairness, its current design requires direct traversal of the decision tree structure. Consequently, the model’s architecture must be accessible, limiting its application to black-box models where the internal tree structure is not exposed. Future work may focus on adapting FairTTTS for other model classes, such as neural networks, or developing techniques that approximate decision paths in models lacking explicit tree structures. Additionally, exploring alternative fairness metrics, refining the probabilistic adjustment mechanisms, and studying the method’s behavior under shifting data distributions could further improve performance and broaden its applicability.

\subsection{Social Impact and Ethical Considerations}

FairTTTS can help mitigate biases affecting marginalized communities by improving fairness metrics at inference time, potentially leading to more equitable outcomes in areas like credit, hiring, or healthcare. However, no single technique can eliminate all discrimination, and fairness definitions vary across contexts. Domain expertise, stakeholder input, and regulatory guidance remain crucial when selecting protected attributes, interpreting metric improvements, and addressing intersectional biases. FairTTTS should thus be viewed as one part of a broader strategy that includes ongoing monitoring, community engagement, and adherence to legal standards.

The FairTTTS method offers a principled yet flexible approach to incorporating fairness into machine learning classification tasks. By utilizing Monte Carlo simulations and dynamic adjustments, the method achieves better fairness without sacrificing accuracy, outperforming traditional method and the baseline without bias mitigation, in diverse scenarios. These results affirm the effectiveness of incorporating fairness considerations on the model's outcomes using a Monte Carlo approach. The FairTTTS method stands out as a robust and practical solution for mitigating bias in production ML classifiers, offering tangible benefits over existing fairness-enhancing methods.

\section{Code Availability}
The benchmark datasets and our method's reproducible source code are available at  https://github.com/nuritci/FairTTTS 



\bibliographystyle{ACM-Reference-Format}
\bibliography{ref}

\end{document}